\documentclass{article}

\RequirePackage{snapshot}

\usepackage[final,nonatbib]{nips_2018}

\usepackage[utf8]{inputenc} 
\usepackage[T1]{fontenc}    
\usepackage{url}            
\usepackage{booktabs}       
\usepackage{amsfonts}       
\usepackage{nicefrac}       
\usepackage{microtype}      

\usepackage[utf8]{inputenc} 
\usepackage[T1]{fontenc}    
\usepackage{url}            
\usepackage{amsfonts}       
\usepackage{microtype}      
\usepackage{array}

\usepackage{graphicx}
\usepackage{color}
\usepackage[usenames,dvipsnames,table]{xcolor}
\usepackage{amsmath}
\usepackage{bm}
\usepackage{algorithm}
\usepackage{algorithmicx}
\usepackage{algpseudocode}

\usepackage[pdftex,pdfpagelabels=false]{hyperref}
\hypersetup{
    colorlinks=true,
    linkcolor=Maroon,
    filecolor=magenta,
    citecolor=cyan,
    urlcolor=cyan,
}
\usepackage{subcaption}
\usepackage{cuted}

\usepackage{authblk}

\renewcommand{\eqref}[1]{(\ref{eq:#1})}
\newcommand{\secref}[1]{\S\ref{sec:#1}}
\newcommand{\figref}[1]{Fig.~\ref{fig:#1}}
\newcommand{\tabref}[1]{Table~\ref{tab:#1}}

\newcommand{\webcars}{\url{https://selfdrivingcars.mit.edu/deeptraffic}}
\usepackage{xcolor}

\title{DeepTraffic: Crowdsourced Hyperparameter Tuning of Deep Reinforcement Learning Systems for Multi-Agent Dense
 Traffic Navigation\vspace{0.1in}}

\author{Lex Fridman\thanks{Corresponding author. Email: \texttt{fridman@mit.edu}. Website: \url{https://hcai.mit.edu}.}}
\author{Jack Terwilliger}
\author{Benedikt Jenik}
\affil{Massachusetts Institute of Technology (MIT)}

\begin{document}

\maketitle

\begin{abstract}
  We present a traffic simulation named DeepTraffic where the planning systems for a subset of the vehicles are handled
  by a neural network as part of a model-free, off-policy reinforcement learning process. The primary goal of
  DeepTraffic is to make the hands-on study of deep reinforcement learning accessible to thousands of students,
  educators, and researchers in order to inspire and fuel the exploration and evaluation of deep Q-learning network
  variants and hyperparameter configurations through large-scale, open competition. This paper investigates the
  crowd-sourced hyperparameter tuning of the policy network that resulted from the first iteration of the DeepTraffic
  competition where thousands of participants actively searched through the hyperparameter space.
\end{abstract}

\section{Introduction}




DeepTraffic is more than a simulation, it is a competition that served and continues to serve as a central project for a
course on Deep Learning for Self-Driving Cars. The competition has received over 24 thousand submissions on
\webcars. Our paper considers how this competition served as a means to crowdsource DQN hyperparameters by exploring the
viability of competitor submissions to successfully train agents.  Such an exploration raises practical questions, in a
context of a specific simulated world, about what works and what doesn't for using deep reinforcement learning to
optimize the actions of an agent in that world.

The latest statistics on the number of submissions and the extent of crowdsourced network training and simulation are as
follows:

\begin{itemize}
\item Number of submissions: 24,013 
\item Total network parameters optimized: 572.2 million
\item Total duration of RL simulations: 96.6 years
\end{itemize}

Deep reinforcement learning has shown promise as a method to successfully operate in simulated physics environments like
MuJoCo \cite{todorov2012mujoco}, in gaming environments \cite{bellemare2013arcade,mnih2013playing}, and driving
environments \cite{shalev2016long,shalev2016safe}. Yet, the question of how so much can be learned from such sparse
supervision is not yet well explored. We hope to take steps toward such understanding by drawing insights from the
exploration of crowdsourced hyperparameter tuning (as discussed in \secref{exploration}) for a well-defined traffic
simulation environment. This includes analysis of network size, temporal dynamics, discounting of reward, and impact of
greedy behavior on the stability and performance of the macro-traffic system as a whole.

\begin{figure*}[ht!]
  \centering
  \includegraphics[width=1.0\textwidth]{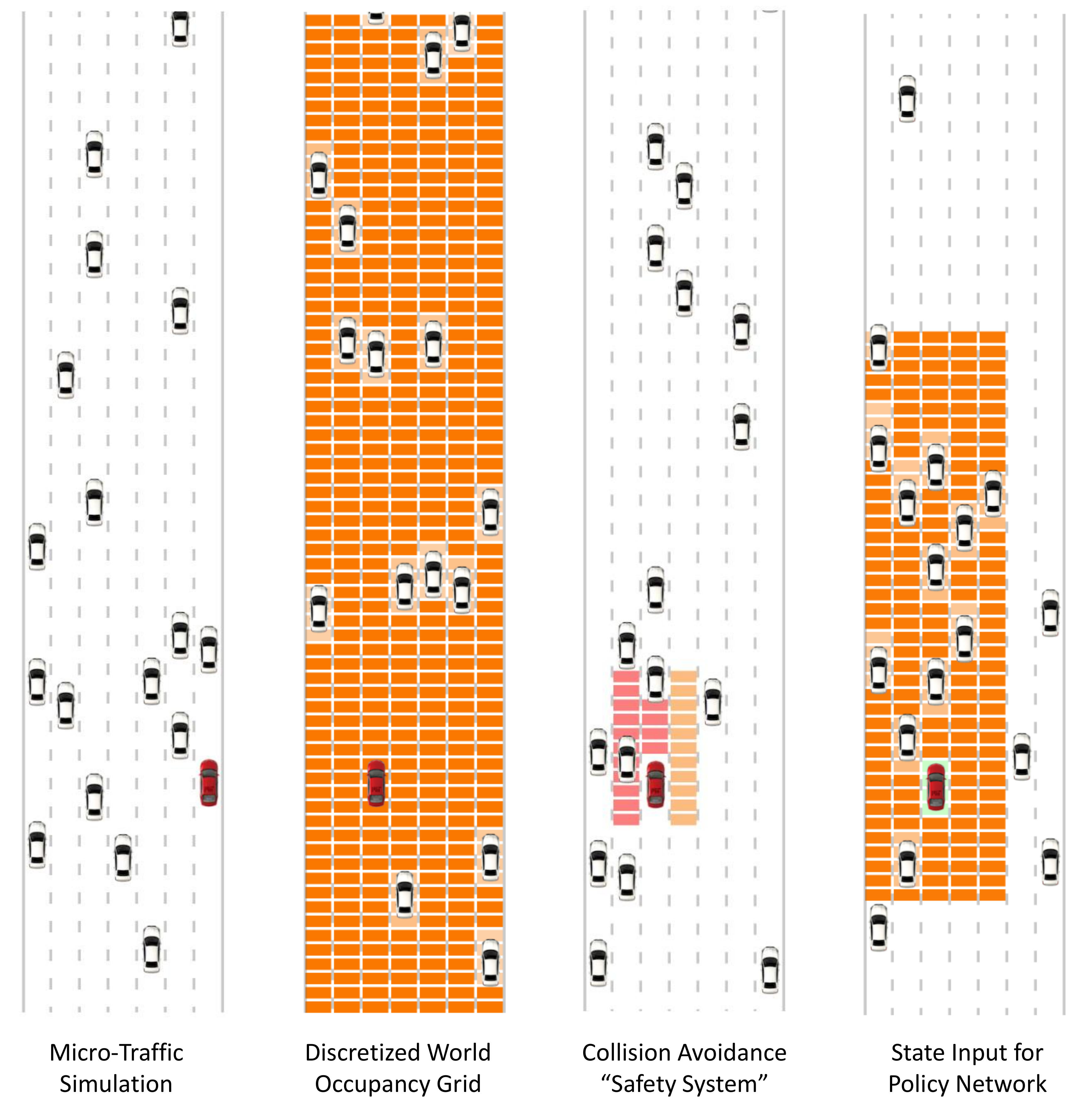}
  \caption{Four perspectives on the DeepTraffic environment: the simulation, the occupancy grid, the collision avoidance
    system, and the slice of the occupancy grid that represents the reinforcement learning ``state'' based on which
    the policy network learns to estimate the expected reward received by taking each of the five available actions. }
  \label{fig:game}
\end{figure*}

The central goals and contributions of DeepTraffic are:

\begin{itemize}
\item \textbf{Exploration and Education}: Provide an accessible, frictionless tool to explore deep reinforcement
  learning concepts for both complete beginners and advanced researchers in reinforcement learning. The goal of
  DeepTraffic, in this aspect, is for competitors to gain understanding, intuition, and insights of how these methods
  can be tuned to solve a real-world problem (i.e., behavioral layer of autonomous vehicles movement planning).
\item \textbf{How Humans Perform Hyperparameter Tuning}: Provide observations on how thousands of humans explore
  hyperparameters of a ``black box'' machine learning system.
\item \textbf{Autonomous Vehicles in Heterogeneous Traffic Simulation}: Provide a methodology for studying impact
  of autonomous vehicles in traffic environments that involve both manually and autonomously controlled vehicles.
\end{itemize}

\begin{figure*}[ht!]
  \centering
  \includegraphics[width=\textwidth]{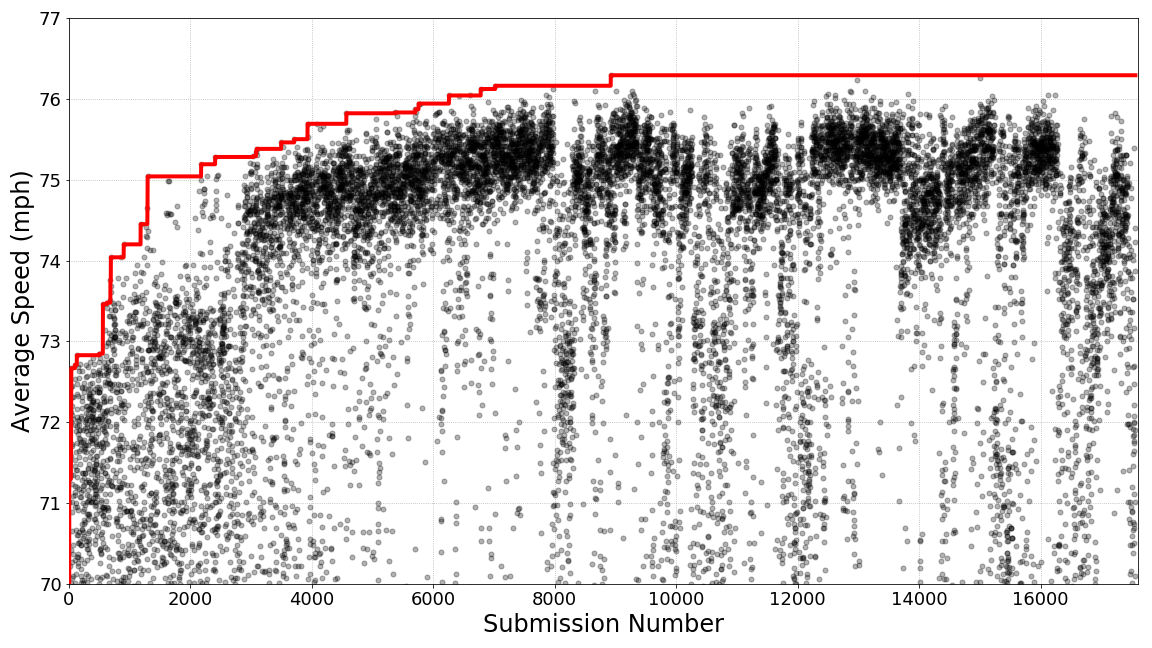}
  \caption{A scatter plot of submission scores. Each black circle corresponds to a DeepTraffic submission. The red
    line shows how the top score increased over time.}
  \label{fig:score}
\end{figure*}

\section{Related Work}

\subsection{Deep Reinforcement Learning Methods}

In recent years, deep reinforcement learning, when trained on sufficiently large datasets of experiences, has been used
to surpass human level performance across a variety of discreet action games \cite{mnih2013playing}
\cite{mnih2015human}, widening the previously narrow domains RL had been successful in \cite{tesauro1995temporal}
\cite{riedmiller2009reinforcement} \cite{diuk2008object}. These methods use variants of \cite{van2016deep} deep Q
networks, an extension of \cite{watkins1992q}, which learn to predict the future discounted returns of state-action
tuples.

Outside of Atari-like game environments, which have relatively small action spaces, deep reinforcement learning has been
used successfully in (1) simulated physics environments like \cite{todorov2012mujoco} \cite{lillicrap2015continuous} as
well as in (2) driving specific applications \cite{shalev2016safe}.

\subsection{Deep Reinforcement Learning Benchmarks and Competitions}

A notable feature of the machine learning community is the popularity of competitions, which exist both as benchmarks on
popular datasets \cite{russakovsky2015imagenet}, \cite{deng2009imagenet}, \cite{lecun1998gradient} and as formally
hosted events. They are so popular that hosting such competitions as a means to crowdsource problem solutions has even
become a feasible business model, e.g. Kaggle. However, reinforcement learning competitions present a set of challenges
not present in supervised learning competitions \cite{whitesonreinforcement}. Rather than evaluation being a passive,
static process in which there is a test set of input/outputs, in RL evaluation is an active, dynamic process, and often
stochastic process, in which an algorithm interacts with an environment. When designing an RL competition, there are
several choices to make when deciding on an evaluation metric: (1) whether to measure performance online (while
learning) or offline (after learning) and (2) whether to measure the accumulated reward or a performance metric. Further,
practical considerations, for large competitions include how to easily and safely run a competitor's, potentially
hazardous, code and how to minimize the number of trials needed to fairly compare submissions.

\subsection{Hyperparameter Tuning}

As in other machine learning domains, hyperparameter tuning is an important component of reinforcement learning. There
are several popular types methods for tuning the hyperparameters of neural networks: grid search, random search
\cite{bergstra2012random}, simulated annealing \cite{kirkpatrick1983optimization}, bandit-optimization
\cite{li2016hyperband} \cite{srinivas2009gaussian}, and black-box Bayesian Optimization methods \cite{golovin2017google}
\cite{shahriari2016taking} \cite{snoek2012practical}, as well as manual tuning. In this paper, we look at crowdsourcing
as a means for hyperparameter tuning.

\begin{figure*}[!ht]
  \centering
  \begin{subfigure}[t]{\textwidth}
    \centering
    \includegraphics[width=\textwidth]{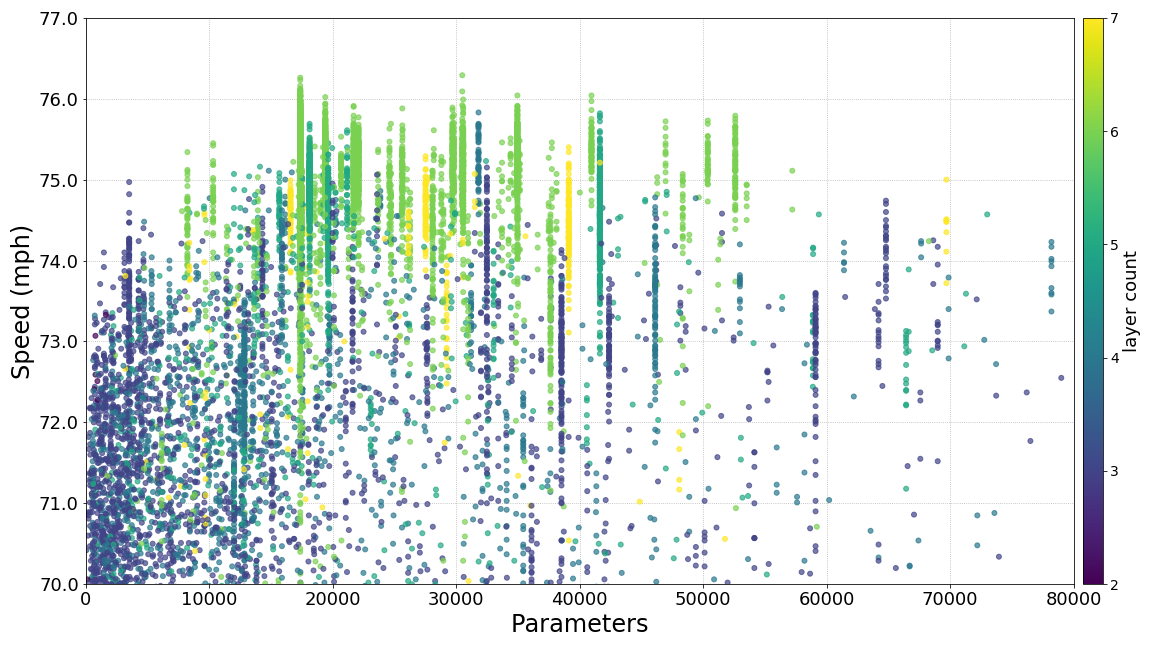}
    \caption{Circle color indicates number of hidden layers from 1 layer (blue) to 7 layers (yellow).}
    \label{fig:params-vs-layers}
  \end{subfigure}\\\vspace{0.1in}
  \begin{subfigure}[t]{\textwidth}
    \centering
    \includegraphics[width=\textwidth]{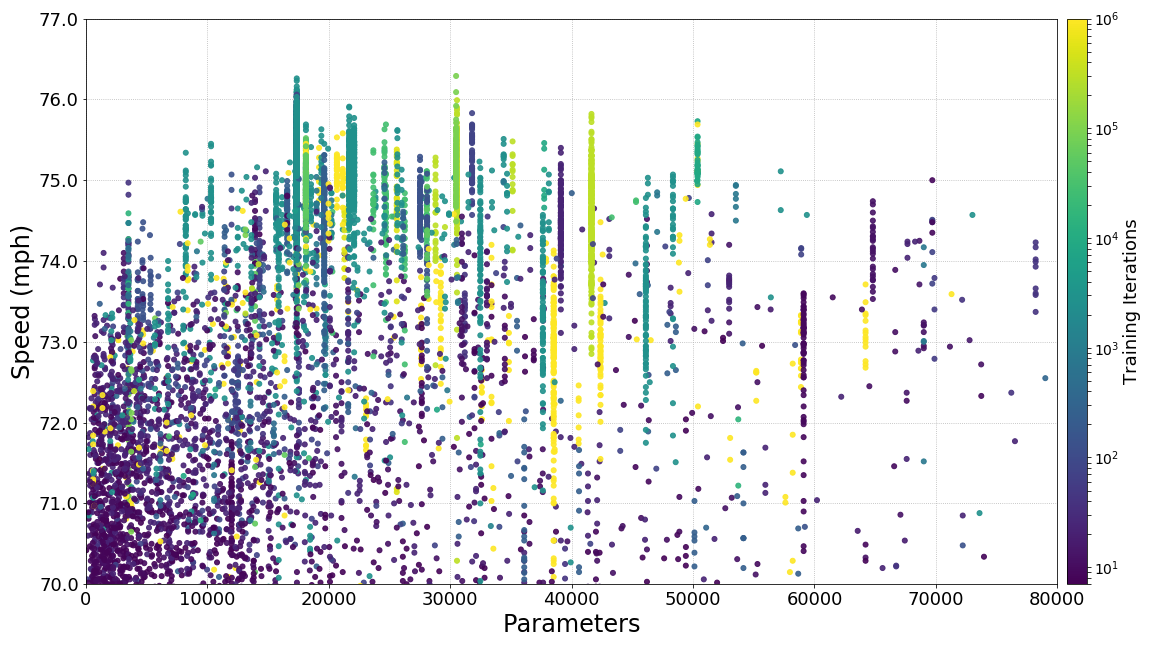}
    \caption{Circle color indicates number of training iterations from 0 (blue) to $10^6$ (yellow).}
    \label{fig:params-vs-train-iterations}
  \end{subfigure}
  \caption{Scatter plots of the number of learnable parameters of a submission and submission score. Submissions with
    scores below 70 mph are not shown.}
  \label{fig:params}
\end{figure*}

\section{DeepTraffic Simulation and Competition}

DeepTraffic is a simulation and deep reinforcement learning environment in which one or more vehicles (red cars) must
navigate through dense traffic (white cars) as quickly as possible. The red cars are ``intelligent'' in that they are
controlled by Javascript code that is provided through the online interface, and can thus plan movement in response to
current road conditions as perceived through its sensors. The white cars move randomly according to a stochastic model
discussed in \secref{simulation}. There is always one red \textit{ego vehicle}, an agent the game screen follows.

\subsection{Simulation}\label{sec:simulation}



\subsubsection{Game Environment}

DeepTraffic takes place on a 7 lane highway. The highway, $H$, is a continuous 2d space, 140 pixels wide by 700 pixels
long. The perspective of the game follows the ego vehicle such that the nose of the ego vehicle remains at the
$175^{th}$ pixel. This means, the dynamics of DeepTraffic are relative to the ego vehicle.

When other vehicles drive off the highway, either by moving slower or faster
than the ego vehicle, they will reappear at the end of the highway opposite to
where they recently disappeared. When vehicle's reappear, their speed and lane
are chosen randomly.

The problem of perceiving the environment is abstracted away by using an
occupancy grid, $H'$ (see second column of \figref{game}). A cell $h' \in H'$
takes on the speed of a vehicle which occupies it:

\newcommand{\speed}{\mathbf{speed}}

\begin{equation*}
  h' = \begin{cases}
    \speed_i & \text{if } \exists c_i \in C : \mathbf{occupies}(h', (x_i, y_i))\\
    0 &\text{otherwise}
  \end{cases}
\end{equation*}

\noindent where $\mathbf{occupies}$ indicates whether any points of a vehicle $c_i$ occupies the
space of $h'$. 

\subsubsection{Vehicle Dynamics}

There are 20 vehicles in DeepTraffic. Each vehicle $c_i$ (1) occupies a space 20
pixels wide by 40 pixels long, (2) is positioned on the highway at $(x_i, y_i)$,
(3) has a top speed of $\speed_{i, max}$, (4) travels at a fraction of its top
speed where the fraction is noted as $\speed_{i,\text{factor}}$, and (5) takes actions
when $t mod t_i = 0$, where $t$ is the time in the game and $t_i$ controls when
$c_i$ makes actions.

The dynamics of the cars in DeepTraffic are all relative to the ego car, $c_0$.
According to these dynamics, a car may move vertically according the equation:

$$\frac{dy}{dt} = -(\speed_i - \speed_0)$$ where $\speed_j=\speed_{j, max} *
\speed_{j, \text{factor}}$.

A car may also move horizontally to its desired target lane according to the equation:

$$\frac{dx}{dt} = (x_{\text{target}} - x_i)/T$$ where $T$ is a time constant which is
equal to the decision cycle.

\subsubsection{Safety System}

Since the focus of DeepTraffic is to learn efficient movement patterns in
traffic, the problem of collision avoidance is abstracted away by using a
``safety system''. The safety system looks at the occupancy grid and prevents
the vehicle from colliding by either (1) preventing an action which would lead
to a crash or (2) altering the speed of the vehicle.

If a vehicle is precisely 4 cells below another vehicle, it assumes the above
vehicle's speed. If the vehicle is closer than 4 cells to an above vehicle, as
can happen after lane changes, it slows down inversely proportional to the
distance. Otherwise, vehicle speeds are unaffected. Formally the safety system
is defined as:

\newcommand{\distance}{\mathbf{d}}

\begin{equation*}
  \speed_i =
    \begin{cases}
      \speed_j     & \text{if }       \distance_{\text{long}}(i, j) = 4 \\
      \speed_j / 2 & \text{if }       \distance_{\text{long}}(i, j) < 4 \\
      \speed_i     & \text{otherwise}
    \end{cases}
\end{equation*}

                    

where the vertical distance from i to j is: $$\distance_{\text{long}}(i, j) = (y_j - y_i)/10$$

The safety system can also override the vehicle's ability to change lanes. If a
vehicle is in any of the 4 occupancy grids immediately to the right (left) of the deciding
vehicle, the 1 occupancy grid behind and to the right (left) of the deciding
vehicle, or in any of the 6 grids ahead and to the right (left) of the deciding
vehicle, the deciding vehicle will remain in its current lane.

\subsubsection{Initialization}

At the beginning of an episode of DeepTraffic, the initial locations of the
vehicles is randomized such that no vehicle is in range of any other vehicle's
safety system. Red car speeds are set to 80 mph. The white car speeds are set to
65 mph.

The state of DeepTraffic can be expressed formally as follows:
\begin{itemize}
  \item $t$: the current frame of the game.
  \item $H$: the 140x700 pixel highway.
  \item $H'$: the 7x70 cell occupancy grid.
  \item $C$: the set of cars.
  \item $t_i$: the decision frequency of car $c_i$. Car $c_i$ chooses an action when
    $t\: mod\: t_i = 0$.
  \item $\speed_{i, \text{factor}}$: the gas pedal $i$.
  \item $\speed_{i, max}$: the top speed of car $i$.
  \item $(x_i, y_i)$: the location of car $i$.
  \item $l_i$: the target lane of car $i$.
\end{itemize}

\begin{table}[h!]
\centering
\begin{tabular}{|>{\raggedright\arraybackslash}p{2.6cm}%
  |>{\centering\arraybackslash}p{1.1cm}%
  |>{\raggedright\arraybackslash}p{4.2cm}%
  |>{\raggedright\arraybackslash}p{5cm}|%
  }
\hline
\textbf{Name}                  & \textbf{Symbol}     & \textbf{DeepTraffic Variable Name}              & \textbf{Description}                                                                                                                          \\ \hline
Patches Ahead                  &            & patchesAhead                      & The number of cells ahead an agent senses.                                                                                           \\ \hline
Patches Behind                 &            & patchesBehind                     & The number of cells behind an agent senses.                                                                                          \\ \hline
Lanes to the Side              &            & lanesSide                         & The number of lanes to each side an agent senses.                                                                                    \\ \hline
Temporal Window                &            & temporal\_window                  & The number of previous states an agent remembers.                                                                                    \\ \hline
Other Autonomous Agents        &            & other\_agents                     & The number of non-ego vehicle agents controlled by a clone of a competitor's code.                                                   \\ \hline
Momentum                       &            & tdtrainer\_options.momentum       & Gradient descent momentum.                                                                                                           \\ \hline
Learning Rate                  & $\alpha$   & tdtrainer\_options.learning\_rate & Gradient descent learning rate.                                                                                                      \\ \hline
l2-regularization              & $l^2$ & tdtrainer\_options.l2\_decay      & The l2 regularization applied to the weights of the neural network. This keeps the weights of the network small.                     \\ \hline
Batch Size                     &            & tdtrainer\_options.batch\_size    & Batch size for gradient descent.                                                                                                     \\ \hline
Discount Factor                & $\gamma$   & opt.gamma                         & Future discount for reward, e.g. $r_0 + \gamma^1 r_1 + ... + \gamma^t r_t$                                                               \\ \hline
Experience Size                &            & opt.experience\_size              & The number of examples in replay memory, i.e. the number of previous experiences to randomly sample from.                            \\ \hline
Epsilon Min                    &            & opt.epsilon\_min                  & Epsilon is the probability that an agent takes a random action. The lowest value epsilon will take on during training.               \\ \hline
Epsilon Test Total             &            & opt.epsilon\_test\_time           & The value of epsilon used outside evaluation.                                                                                        \\ \hline
Total Learning Steps           &            & opt.learning\_steps\_total        & The total number of steps the agent will learn for. (Hopefully you set this to the number of training iterations)                    \\ \hline
Learning Threshold             &            & opt.start\_learning\_threshold    & The minimum number of examples in experience replay memory before the agent begins the learning process.                             \\ \hline
Learning Steps Before Burnin   &            & opt.learning\_steps\_burning      & The number of random actions the agent takes before beginning the learning process. This is to collect a large number of experiences \\ \hline
\end{tabular}
\vspace{0.1in}
\caption{This table summarizes the hyperparameters in DeepTraffic.}
\label{tab:parameters}
\end{table}

\subsection{Implementation}\label{sec:implementation}

One of the defining qualities of DeepTraffic is that it is entirely implemented in Javascript, including the simulation,
the visualization, the reinforcement learning framework, the neural network training and inference. This significantly
simplifies the typical overhead of installing software required to train a Deep RL agent and run a simulated
environment. The accessibility of this environment as a way to explore Deep RL approaches naturally motivated turning
the playground into a competition (see \secref{competition}), and consequently an education tool. Competitors
participate by submitting Javascript code that controls the red vehicle. This code is usually one that utilizes a DQN.


The browser is a non-traditional platform to deploy a neural network. However, the browser is a ubiquitous platform, and
running someone's code is often as trivial as visiting a web page. We chose ConvNetJS since it uses plain Javascript,
rather than many other frameworks which rely on WebGL, which is not universally supported.

Despite its unconventionality, ConvNetJS bears resemblance to other deep learning frameworks. Data, weights, and
gradients are stored in multi-dimensional arrays.  Optimizers such as SGD can be used to train neural networks.


The competition's default reinforcement learning algorithm was Deep Q-learning with Experience Replay, as it is detailed
in \cite{mnih2013playing}. Nearly all competitors used this default implementation. Briefly, Deep Q-learning uses a
neural network to approximate an action-value function, $Q(s, a)$, which maps action-state pairs to the ``quality'' or
expected future discounted return.

The network uses the following gradient to learn the $Q$ function:
$$\nabla L(\theta) = \mathbb{E}\bigg[ \bigg( \overbrace{r_t +
  \gamma\max_{a_t}Q(s_{t+1}, a_{t+1}; \theta)}^{\text{Bellman updated return}} -
\overbrace{Q(s_t, a_t; \theta)}^{\text{predicted return}} \bigg)\nabla_{\theta_i} Q(s,a;\theta_i)  \bigg]$$
where, during each training step, many observations are sampled from replay memory.

 \begin{figure}[ht!]
   \centering
   \includegraphics[width=\textwidth]{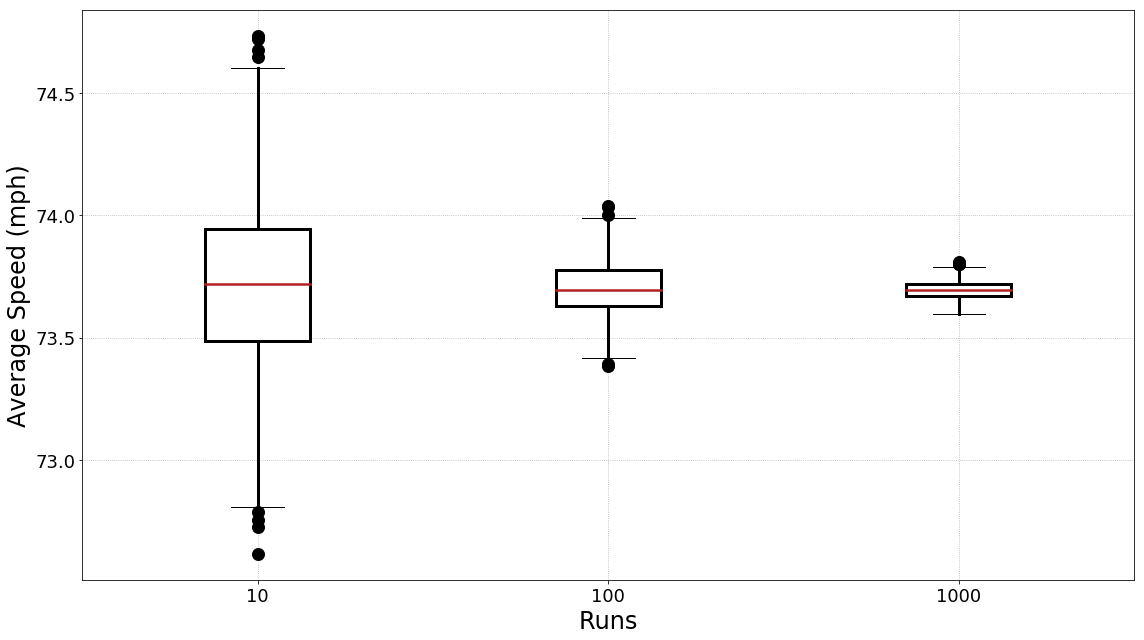}
   \caption{A box plot showing that submission scores become more robust to the
     stochastic nature of the game. As more runs are used to in the evaluation
     score. Each run includes 10,000 simulation time steps. A submission score
     is calculated by taking the median score over the runs.}
   \label{fig:evaluation}
 \end{figure}

 \begin{figure}[ht!]
   \centering
   \includegraphics[width=\textwidth]{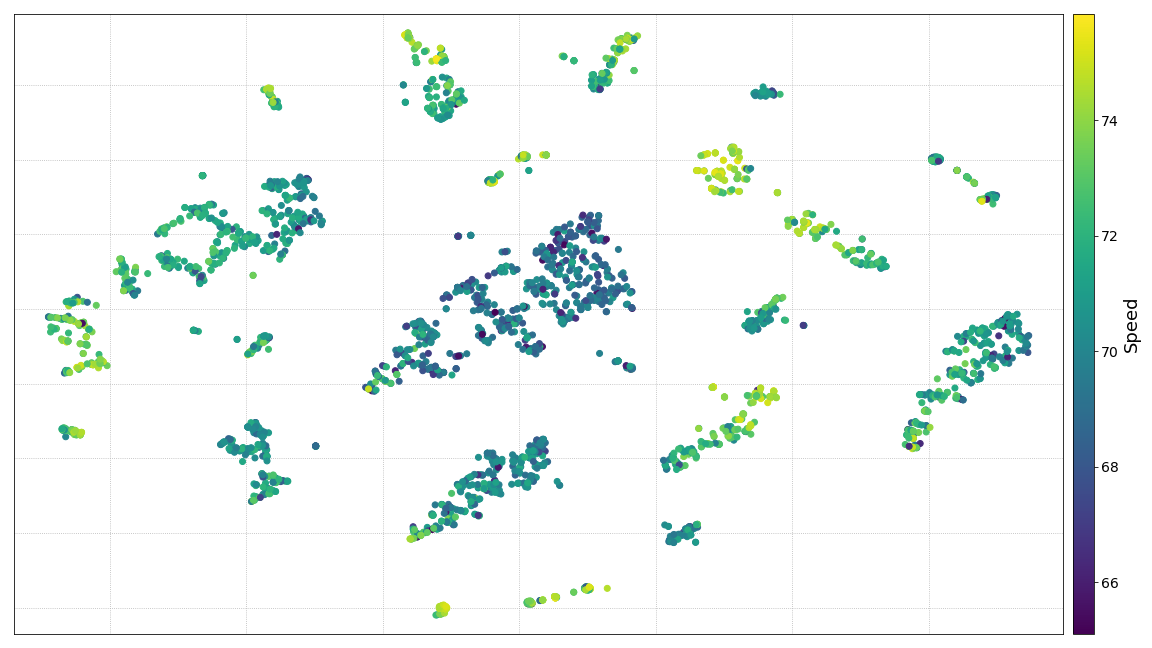}
   \caption{A visualization of unique DeepTraffic submissions produced with t-SNE. The higher-dimensional representation
     used to generate this plot defined submissions as vectors over the following variables: patches ahead, patches
     behind, l2 decay, layer count, gamma, learning rate, lanes side, training iterations. Circle color corresponds to
     submission score.}
   \label{fig:evaluation}
 \end{figure}

\subsection{Competition}\label{sec:competition}

Participants in the competition are provided with a default DQN implementation
\cite{karpathy2014convnetjs}, where they get to configure different
hyperparameters to achieve the best performance, where performance is measured
as the average speed of all the red cars. Competitors are able to change the
size of the DQN's input by defining the number of patches/cells ahead of the
front of the car, behind it and next to the car, the network gets to look at. In
addition, they can specify the network layout, meaning adding and removing
layers, and changing their sizes and activations. Beyond that, participants can
also configure training options like learning rate and regularization methods,
and also reinforcement learning specific parameters like exploration vs
exploitation and future reward discounts.

After a competitor has selected some hyperparameters they press a button to train the network in the browser, using a
separate thread, referred to as a web-worker. While training, competitors can see improvements running live in the main
visualization, as the DQN running on the main thread is periodically replaced by the most recent version training in the
web-worker. When the participants are satisfied with the quality of their network they can submit the network for
server-side evaluation to earn a spot in the leaderboard.

An interesting observation in the competitive chase toward greater performing Deep RL agents that is common to many Deep
RL tasks is that neither the authors nor the competitors had an explicit model for an optimal policy. The high
dimensionality and the stochastic nature of the state space made it intractable for model-based path planning methods
\cite{karaman2011anytime}. In the early days of the competition, many competitors claimed that it's impossible for the
car to achieve a speed above 73 mph. Once that barrier was broken, 74 mph became the new barrier, and so on, reaching
the current socially-defined performance ceiling of 76 mph. From the social perspective,
 it is interesting to observe that such self-imposed ceilings often led to performance plateaus, much like those in
 other competitive endeavors \cite{bannister2004four}.

Network parameters were not shared publicly between competitors, but in several cases, small online communities formed to
distribute the hyperparameter tuning process across the members of that community. An interesting result of such
distributed hyperparameter tuning efforts was that occasional plateaus in performance were broken by one individual and
then quickly matched and superseded by others.

\subsection{Evaluation Process}
Robust evaluation is an important aspect of any competition. When constructing
the evaluation process for DeepTraffic we identified the following challenge:
design an evaluation process in which (1) score variance is kept to a minimum
with reasonable computing resources and (2) arbitrary Javascript code can be run
safely. Unlike many RL competitions, DeepTraffic is not scored using accumulated
reward during online learning, rather, it is scored by taking the median
average-speed over several simulation runs.

The primarily challenge in designing for (1) is that the nature of DeepTraffic
is stochastic, and therefore any finite evaluation will be based on only the
starting conditions sampled. One way to approach this problem is to design a
score which aggregates over a sample of starting conditions. Evaluation variance
is then governed by the run length and the number of runs per evaluation. An
additional parameter controlling the score variance is the method which
initializes the game prior to each run.

The primary challenge in designing for (2) is that users may submit arbitrary
code to DeepTraffic and therefore (a) the program defining the DeepTraffic
environment must be unalterable and (b) the evaluation process must be secure
from malicious code. Addressing (a), we run submission code on our own server by
loading the environment after user code. Addressing (b), running code in a
web-worker prevents it from performing any harmful IO operations.







\newcommand{\tunehspace}{\hspace{0.4in}}
\begin{figure}[htp!]
  \begin{subfigure}[t]{0.47\textwidth}
    \includegraphics[width=\textwidth]{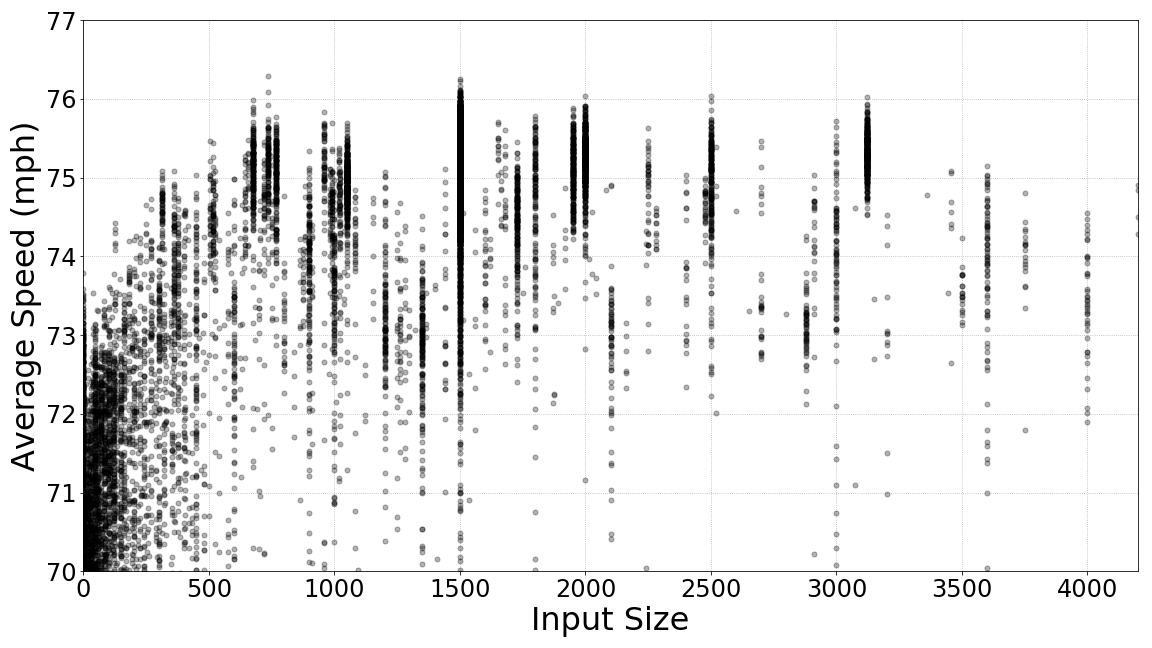}
    \caption{}
    \label{fig:tune1}
  \end{subfigure}\tunehspace
  \begin{subfigure}[t]{0.47\textwidth}
    \includegraphics[width=\textwidth]{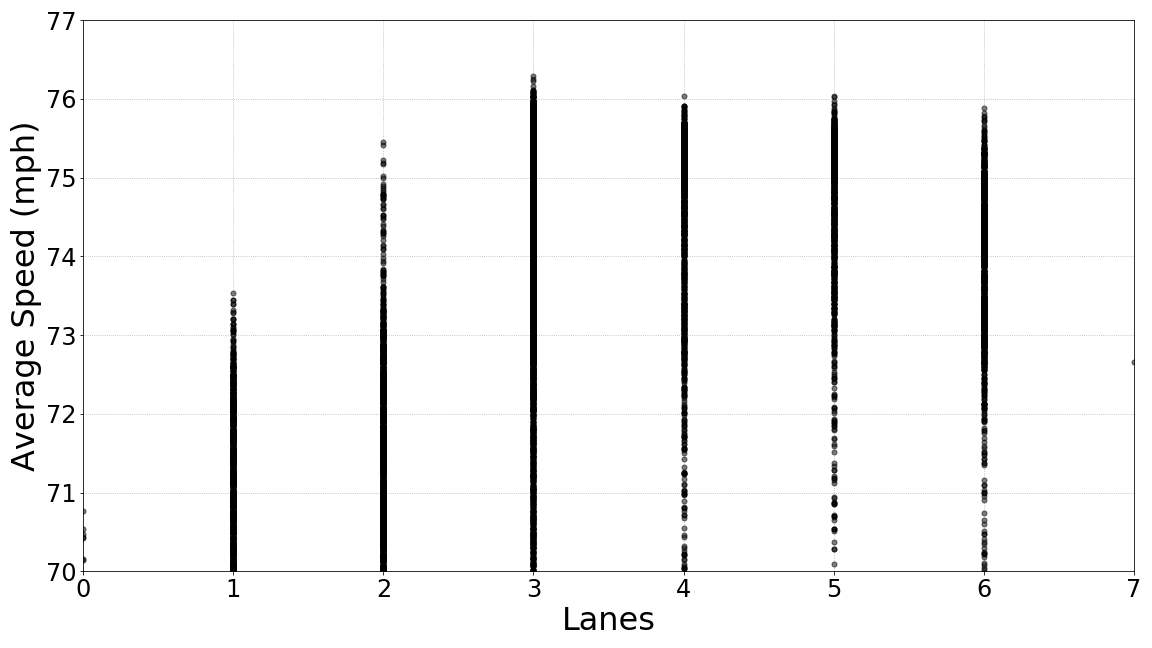}
    \caption{}
    \label{fig:tune2}
  \end{subfigure}\\\vspace{0.2in}
  \begin{subfigure}[t]{0.47\textwidth}
    \includegraphics[width=\textwidth]{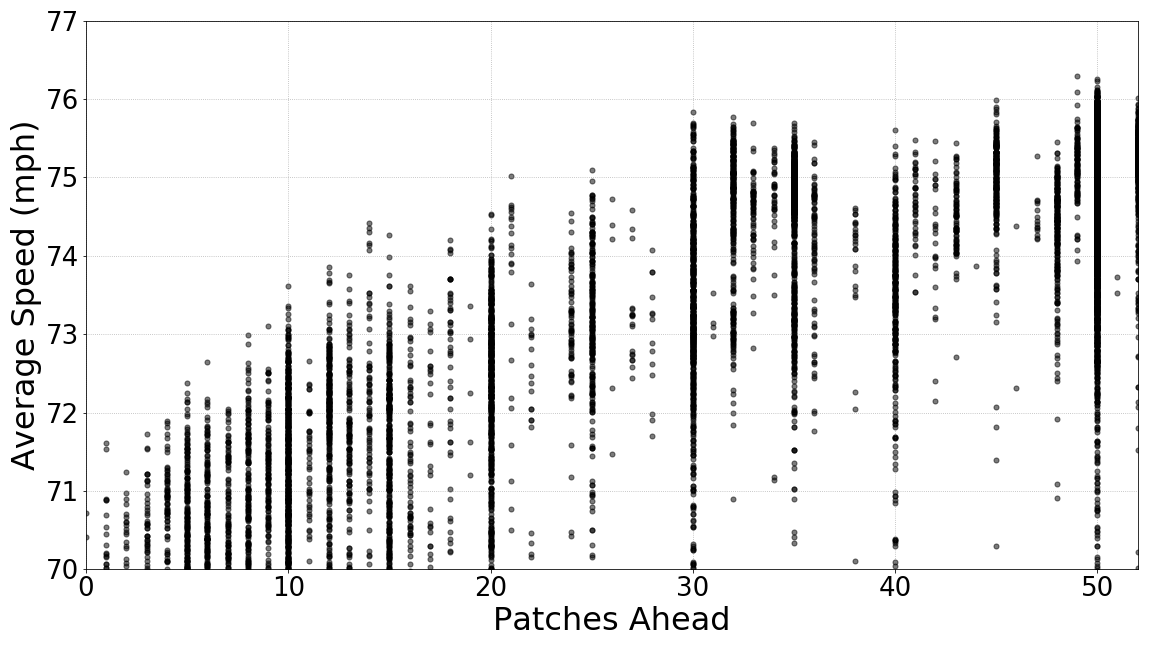}
    \caption{}
    \label{fig:tune3}
  \end{subfigure}\tunehspace
  \begin{subfigure}[t]{0.47\textwidth}
    \includegraphics[width=\textwidth]{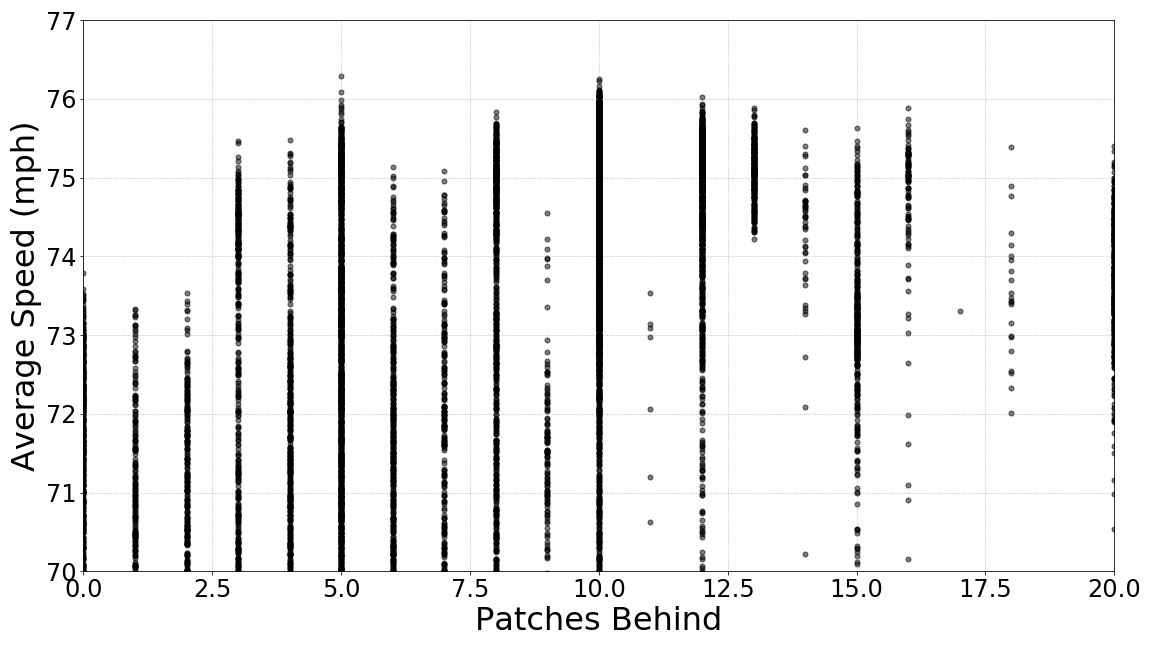}
    \caption{}
    \label{fig:tune4}
  \end{subfigure}\\\vspace{0.2in}
  \begin{subfigure}[t]{0.47\textwidth}
    \includegraphics[width=\textwidth]{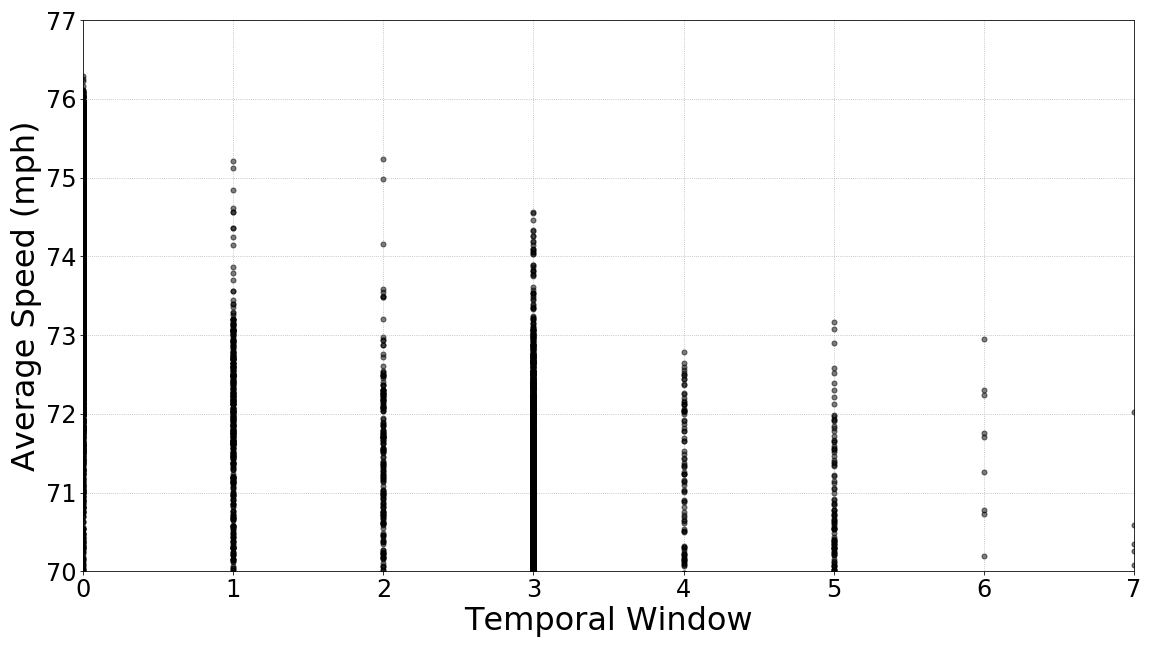}
    \caption{}
    \label{fig:tune5}
  \end{subfigure}\tunehspace
  \begin{subfigure}[t]{0.47\textwidth}
    \includegraphics[width=\textwidth]{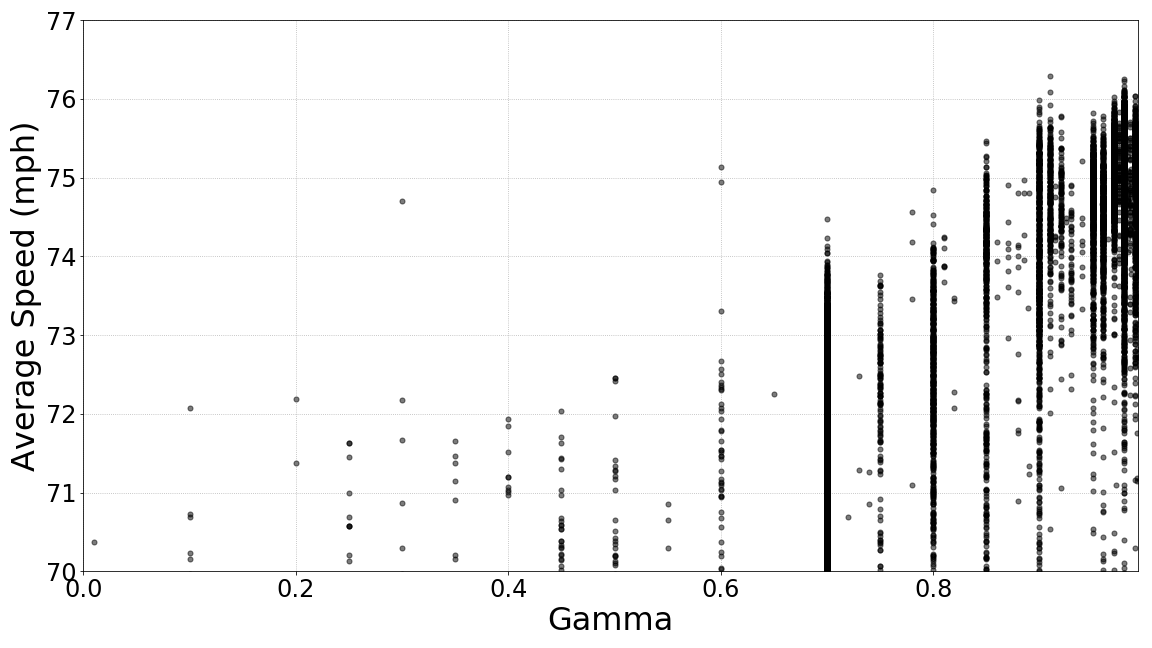}
    \caption{}
    \label{fig:tune6}
  \end{subfigure}
  \begin{subfigure}[t]{0.47\textwidth}
    \includegraphics[width=\textwidth]{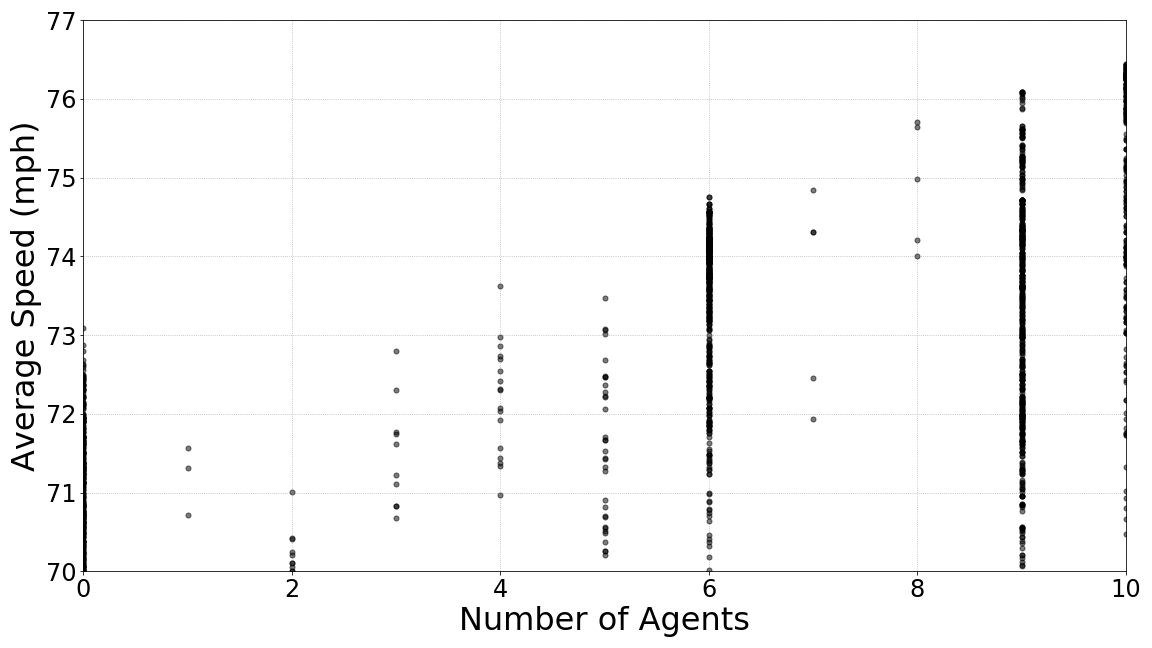}
    \caption{}
    \label{fig:tune7}
  \end{subfigure}
  \caption{Scatter plots comparing submission parameters and submission scores (see \secref{evaluation}).}
  \label{fig:tune}
\end{figure}

\section{Network Performance and Exploration}\label{sec:exploration}

The large-scale crowdsourcing of hyperparameters through over 13,000 network submissions allows us to gather
observations about what hyperparameters have a significant impact on evaluated agent performance and what come at a
great cost without much performance gain. The key insights gathered from this exploration are highlights in the
following subsections. In support of these insights, \figref{score} shows the evolution of submission scores over time,
\figref{params} shows the impact of network size, depth, and training iteration, and \figref{tune} shows a rich set of
impactful hyperparameters as they interact with final network performance in intricate and sometimes counter-intuitive
ways.

\subsection{Hyperparameters}

There are 15 hyperparameters for DeepTraffic's default DQN implementation, some of which are presented in
\tabref{parameters}. These hyperparameters fall into 3 categories: input parameters which specify how the agent senses
the world, gradient descent parameters which control general neural network gradient parameters, and reinforcement
learning specific parameters which control deep q learning specific parameters.

Input parameters define the spatial and temporal features of the agent input.  Increasing the size of the input provides
the agent with more information but also increases the computational cost of learning.

\subsection{Insight: Size Matters}

The best networks generally have the most parameters and many layers. As \figref{params-vs-layers} shows, the
conventional wisdom holds up for DeepTraffic: the larger and deeper the network, the better the performance.  However,
there is some diminishing returns when the input size to the network is too large, as indicated by \figref{tune1}. 
Accompanying the increased performance of networks with more parameters, is the requirement that the number of training
iterations increases as well as shown in \figref{params-vs-train-iterations}. Therefore, the cost of using larger
networks is a longer training time, which perhaps contributes to the observed diminishing returns of adding more parameters.

\subsection{Insight: Live in the Moment}

Knowledge of temporal dynamics does not significantly improve the agent's ability to achieve higher reward. This is a
counter-intuitive observation for a simulation that is fundamentally about planning one's trajectory through space and
time. As \figref{tune5} shows, not looking back in time at all provides the best performance. Put another way, the temporal
dynamics of the driving scene that led up to the current state does not have a significant impact on the evolution of
the environment in the future, and thus optimization of actions through that environment does not need to consider the
past. This is surprising since the high-dimensional state space in Deep RL approaches commonly encode the temporal
dynamics of the scene in the definition of the state space \cite{mnih2016asynchronous,mnih2015human}. This is done in
order to capture a sequence of partial observations that in-sum form a more complete observation of the state
\cite{zhu2017improving}. For DeepTraffic, the agent does not appear to need a larger temporal window (injecting
``short-term memory'' into the state representation) in order to more fully specify the current state.

\subsection{Insight: Look Far Forward}

Looking ahead spatially suffers much less from diminishing returns than does looking behind. As shown in \figref{tune3},
the more of the state space in front of the agent that the network is able to consider, the more successful it is at
avoiding situations that block it in. \figref{tune4}, on the other hand, shows that performance gains level off quickly
after more than 5 patches behind the agent are considered. Spatially, the future holds more promise than the
past. Similarly, \figref{tune2} shows that increases how far to the sides the agent looks suffers from diminishing
returns as well, maxing out at 3 lanes to each side. The value of 3 in \figref{tune2} represents sideways visibility
that covers all lanes when the vehicle is positioned in the middle lane. In general, high-performance agents tend to
prefer the middle lane that allows them the greatest flexibility in longer-term navigation through the vehicles ahead of
it.

\subsection{Insight: Plan for the Future}

One of the defining challenges for reinforcement learning is the \emph{temporal credit assignment problem}
\cite{sutton1998reinforcement}, that is, assigning value to an action in a specific state even though that action's
consequences do not materialize until much later in time. As shown in \figref{tune6}, minimizing the discounting of
future reward by increasing $\gamma$ (referred to as ``gamma'' in the figure) nearly always improves performance. Much
like the previous insight, the future is valuable for estimating the expected reward and planning ones actions
accordingly. Achieving good average speed requires that the agent avoid clusters of other cars, and often the avoidance
strategy requires planning many moves ahead.

\subsection{Insight: Evaluation is Expensive}\label{sec:evaluation}

\figref{evaluation} shows that it takes at least ten million simulation time steps (shown as 100 evaluation runs in the
figure) to converge towards a stable estimation of deep reinforcement learning agent performance with a standard
deviation of possible scores falling below 0.1. One of the open problems of running a deep reinforcement learning
competition is to have an effective way of ranking the performance of the submitted policy networks. The very large size
and non-deterministic nature of the state space make stable, consistent, and fair evaluation of an agent very
challenging, given the amount of computational resources it takes to execute a forward pass 10+ million times through a
network with 40,000+ parameters for each of the 13,000+ agents submitted to date.





\section{Conclusion}

In this work we seek to make deep reinforcement learning accessible to tens of thousands of students, researchers, and
educators by providing a micro-traffic simulation with in-browser neural network training capabilities.
We look back at the crowdsourced hyperparameter space exploration and draw insights from what was effective
to improving overall agent performance. We provide the traces of this exploration as the ``Human-Based Hyperparameter
Tuning'' dataset.

\bibliographystyle{plain}
\bibliography{deeptraffic}

\end{document}